\pdfoutput=1
\documentclass[11pt]{article}

\usepackage[]{ACL2023}

\usepackage{times}
\usepackage{latexsym}

\usepackage{verbatim}
\usepackage{amsmath}
\usepackage{multirow}
\usepackage{booktabs}
\usepackage{tabu}
\usepackage[noend]{algpseudocode}
\usepackage{algorithmicx,algorithm}
\usepackage{pdfpages}
\usepackage{subfigure}
\usepackage{amsfonts,amssymb}
\usepackage{graphicx}

\usepackage[T1]{fontenc}
\usepackage[utf8]{inputenc}

\usepackage{microtype}

\usepackage{inconsolata}

\title{HAHE: Hierarchical Attention for Hyper-Relational Knowledge Graphs in Global and Local Level}

\author{ {\bf Haoran Luo\textsuperscript{\rm 1}, Haihong E\textsuperscript{\rm 1}\thanks{\ \ Corresponding author.}  , Yuhao Yang\textsuperscript{\rm 2}, Yikai Guo\textsuperscript{\rm 3}, Mingzhi Sun\textsuperscript{\rm 1},} \\{\bf Tianyu Yao\textsuperscript{\rm 1}, Zichen Tang\textsuperscript{\rm 1}, Kaiyang Wan\textsuperscript{\rm 1}, Meina Song\textsuperscript{\rm 1}, Wei Lin\textsuperscript{\rm 4}} \\
         \textsuperscript{1}School of Computer Science, Beijing University of Posts and Telecommunications, China \\ 
         \textsuperscript{2}School of Automation Science and Electrical Engineering, Beihang University, China \\ 
         \textsuperscript{3}Beijing Institute of Computer Technology and Application, China \\ 
         \textsuperscript{4}Inspur Group Co., Ltd., China \\ 
         \texttt{\{luohaoran, ehaihong\}@bupt.edu.cn}}

\begin{document}
\maketitle
\begin{abstract}
 Link Prediction on Hyper-relational Knowledge Graphs (HKG) is a worthwhile endeavor. HKG consists of hyper-relational facts (H-Facts), composed of a main triple and several auxiliary attribute-value qualifiers, which can effectively represent factually comprehensive information. The internal structure of HKG can be represented as a hypergraph-based representation globally and a semantic sequence-based representation locally. However, existing research seldom simultaneously models the graphical and sequential structure of HKGs, limiting HKGs' representation. To overcome this limitation, we propose a novel \textbf{H}ierarchical \textbf{A}ttention model for \textbf{H}KG \textbf{E}mbedding (\textbf{HAHE}), including global-level and local-level attention. The global-level attention can model the graphical structure of HKG using hypergraph dual-attention layers, while the local-level attention can learn the sequential structure inside H-Facts via heterogeneous self-attention layers. Experiment results indicate that HAHE achieves state-of-the-art performance in link prediction tasks on HKG standard datasets. In addition, HAHE addresses the issue of HKG multi-position prediction for the first time, increasing the applicability of the HKG link prediction task. Our code is publicly available\footnote{\url{https://github.com/LHRLAB/HAHE}}.
\end{abstract}

\section{Introduction}

Knowledge graphs (KGs) are semantic networks that define entity relationships. Early KG research~\citep{TransE,RotatE,TuckER} use binary relationships, often expressed as a triple-based fact (\textit{subject}, \textit{relation}, \textit{object}). Yet, n-ary relational facts (containing more than two entities) are abundant in real-world KGs like Freebase~\citep{Freebase} and Wikidata~\citep{Wikidata}. \citet{HINGE} represent an n-ary relational fact as a hyper-relational fact (H-Fact) consisting of a main triple(\textit{s},\textit{r},\textit{o}) and several auxiliary attribute-value qualifiers \{(\textit{a$_i$ }:\textit{v$_i$})\}, and KGs composed of H-Facts are called hyper-relational knowledge graphs (HKGs).

\begin{figure}[t]
\centering
\includegraphics[width=7.8cm]{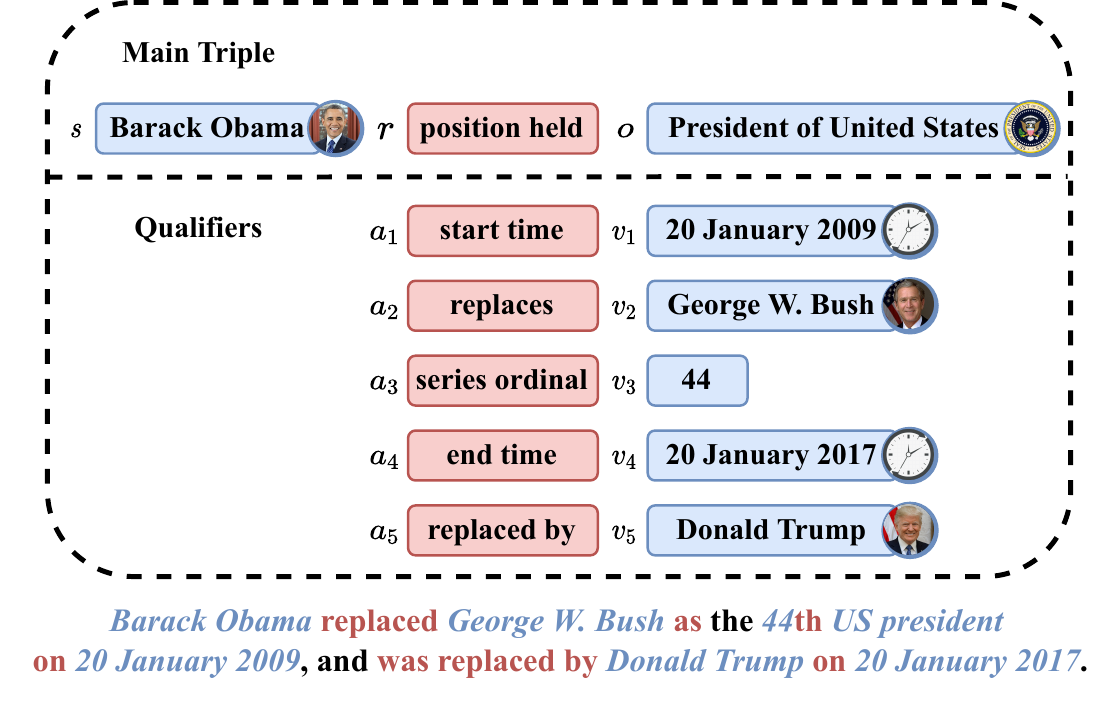}
\caption{An example of hyper-relational fact structure.}
\label{f1}
\end{figure}

\begin{figure*}[h!t]
\centering
\includegraphics[width=16.1cm]{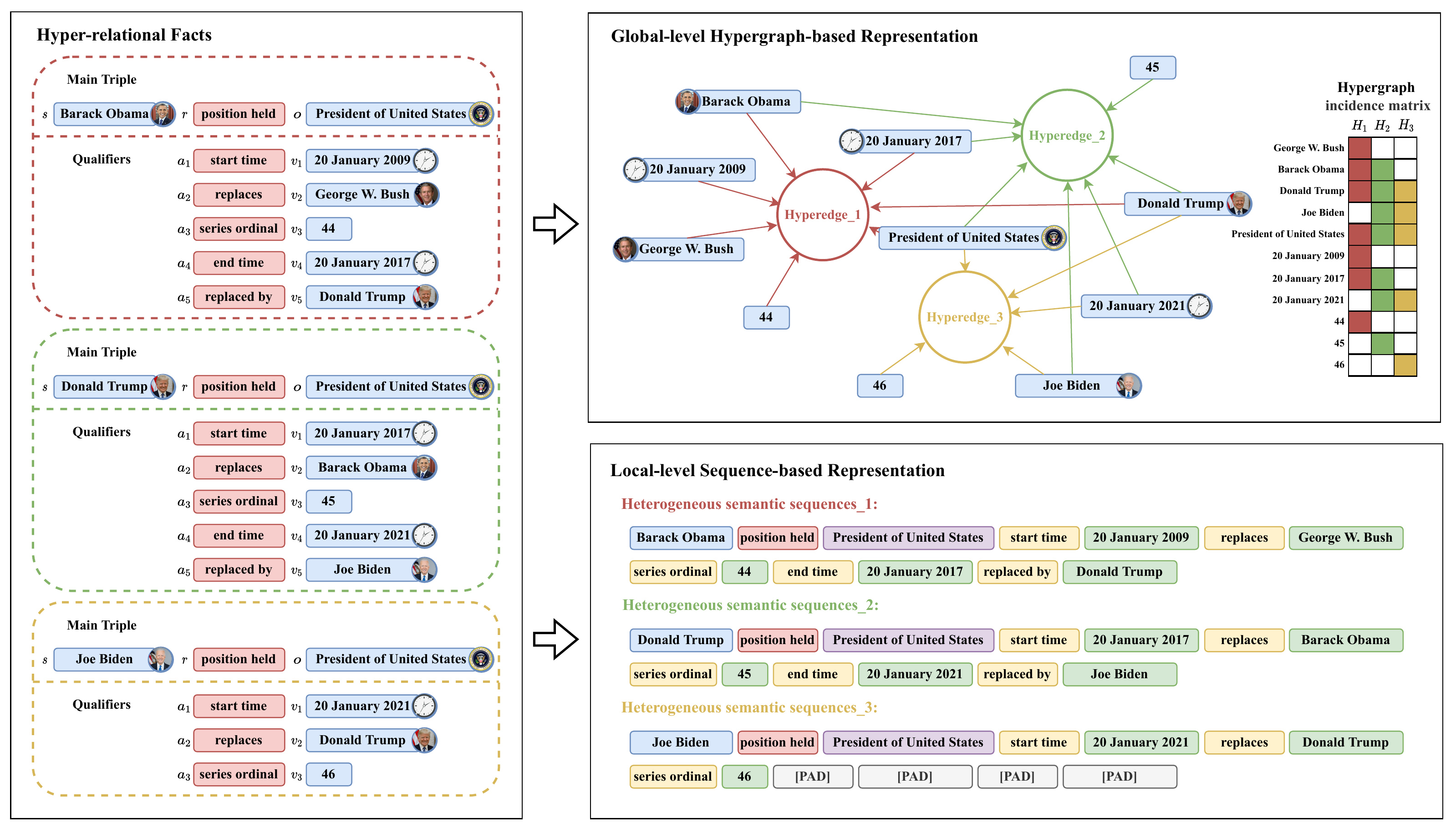}
\caption{The Global-level Hypergraph-based representation and Local-level Sequence-based representation based on three examples of H-Facts in HKGs. }
\label{f2}
\end{figure*}

As shown in Figure~\ref{f1}, an H-fact can describe a real-world fact. Unlike traditional triple-based facts, H-Facts do not just raise the number of entities in facts from two to n. It structurally and effectively represents the n-ary relational facts prevalent in reality. Globally, it extends ordinary graph structure to hypergraph~\citep{Hypergraph} structure. Locally, it defines five heterogeneous roles of \textit{s,r,o,a,v} within facts to capture the semantic information of the fact\textit{`Barack Obama held position as US president'}, as illustrated in Figure~\ref{f2}.

Recent research has demonstrated various embedding strategies for hyper-relational representations. However, current approaches only consider global hypergraph structures or local semantic sequence structures. For instance, StarE~\citep{StarE} employs the information transfer function of graph neural networks (GNN) to unidirectionally pass auxiliary key-value pair information into the main triples' relations, thereby capturing the graph structure but insufficiently between multiple entities and relations within the H-facts. In contrast, GRAN~\citep{GRAN} initially incorporates the Transformer encoder~\citep{Transformer} into the HKG embedding, capturing the fully connected semantic information locally inside H-facts, while disregarding the global structure. Consequently, representing the global and local structure of HKG simultaneously with hierarchical attention becomes a promising research direction, but an inadequate representation of HKG structure constrains HKG embeddings. 

To overcome this limitation, we propose a novel \textbf{H}ierarchical \textbf{A}ttention model for \textbf{H}KG \textbf{E}mbedding (\textbf{HAHE}) that incorporates global-level and local-level attention. We update the global node embeddings using the HKG hypergraph structure. However, by complete connectivity, the previous hypergraph attention network~\citep{HyperGAT} just converts all hypergraph nodes into a regular graph and then utilizes the GAT~\citep{GAT} layer for node embedding updates, rendering it unable to distinguish which nodes comprise a hyperedge. Consequently, we design hypergraph dual-attention layers to aggregate node embedding information into hyper-edge embedding through the attention mechanism. After obtaining the hyper-edge embedding, we update the node embedding by feeding it back to the node through the attention mechanism. In this way, nodes are allowed to learn more distant information from the whole HKG. This hypergraph dual-attention method significantly enhances learning capacity. It then transfers the updated node information to the local level's attention. Inspired by GRAN's heterogeneous attention~\citep{GRAN}, we define five types of nodes and fourteen types of edges in a single H-Fact and develop heterogeneous self-attention layers with both node-bias and edge-bias attention to learn the semantic content of H-Facts. The last step is to output the link prediction findings using an MLP-based decoding process for one-position or multi-position link prediciton tasks on HKGs.

Experiments on link prediction were performed on three HKG standard datasets, JF17K~\citep{m-TransH}, Wikipeople~\citep{NaLP}, and WD50K~\citep{StarE}. The state-of-the-art results indicate that HAHE is effective in the link prediction task. In addition, adequate ablation experiments were designed to highlight the importance of global and local focus, and HAHE is also used for the HKG multi-position prediction task, i.e., predicting two or more entities or relations simultaneously in a single H-fact, hence increasing the applicability of the HKG link prediction task. Ultimately, we make our code publicly available and discuss the limitations and future work of HKG embedding representation. 

% Our most significant contributions are as follows:
% \begin{itemize}
% \item We propose a novel hierarchical attention model for HKGs, to the best of our knowledge, it is the first time to simultaneously model globally graphical representation and locally sequential representation.
% \item We conduct extensive and sufficient experiments to demonstrate that our model achieves state-of-the-art performance for link prediction on HKGs.
% \item To our best knowledge, HAHE is the first model to solve the HKG multi-position prediction task.
% \end{itemize}

\section{Related Work}

Early approaches consider hyper-relational fact(H-Fact) mainly as graph structure, focusing more on the topological relations of entities. For example, m-TransH~\citep{m-TransH} projects the entities onto the relation hyperplane. RAE, NaLP, NeuInfer, and N-TuckER~\citep{RAE,NaLP,NeuInfer,N-TuckER} optimize the method of the Hyper-relational knowledge graph (HKG) embedding based on m-TransH. However, none of them adopts the hyper-relational structure.

HINGE~\citep{HINGE} firstly proposes the attribute-value qualifiers for embedding hyper-relational representation using CNN, and Hyper2~\citep{Hyper2} initializes the relation and entity in the Poincar\'e ball vectors to improve the model accuracy. Yet, neither of these methods considers graphical structure or semantic sequences of HKGs.

StarE~\citep{StarE} employs GNN as the message-passing mechanism to encode entities and relationships, while Transformer is the decoder to get the result. HyTransformer~\cite{HyTransformer} applies layer normalization and dropout methods to replace StarE's encoder. MSeaHKG~\citep{MSeaHKG} proposed that a message-passing function significantly impacts the model performance, so it replaced the static message-passing function in StarE with a dynamic one. These models consider the graph structure within the hyper-relational facts, but disregard the semantic sequences.

GRAN~\citep{GRAN} is an improvement for Transformer~\citep{Transformer}. It replaces Transformer's self-attention with edge-biased fully-connected attention and accurately collects semantic information. Despite this, it ignores the graph structure.

Unlike earlier models, HAHE considers graph structure and semantic sequences simultaneously and employs hierarchical attention for link prediction on HKGs. Moreover, it is the first to improve previous methods by modeling the structure of HKG via global-level embedding and can perform multi-position prediction.

\begin{figure*}[h!t]
\centering
\includegraphics[width=16cm]{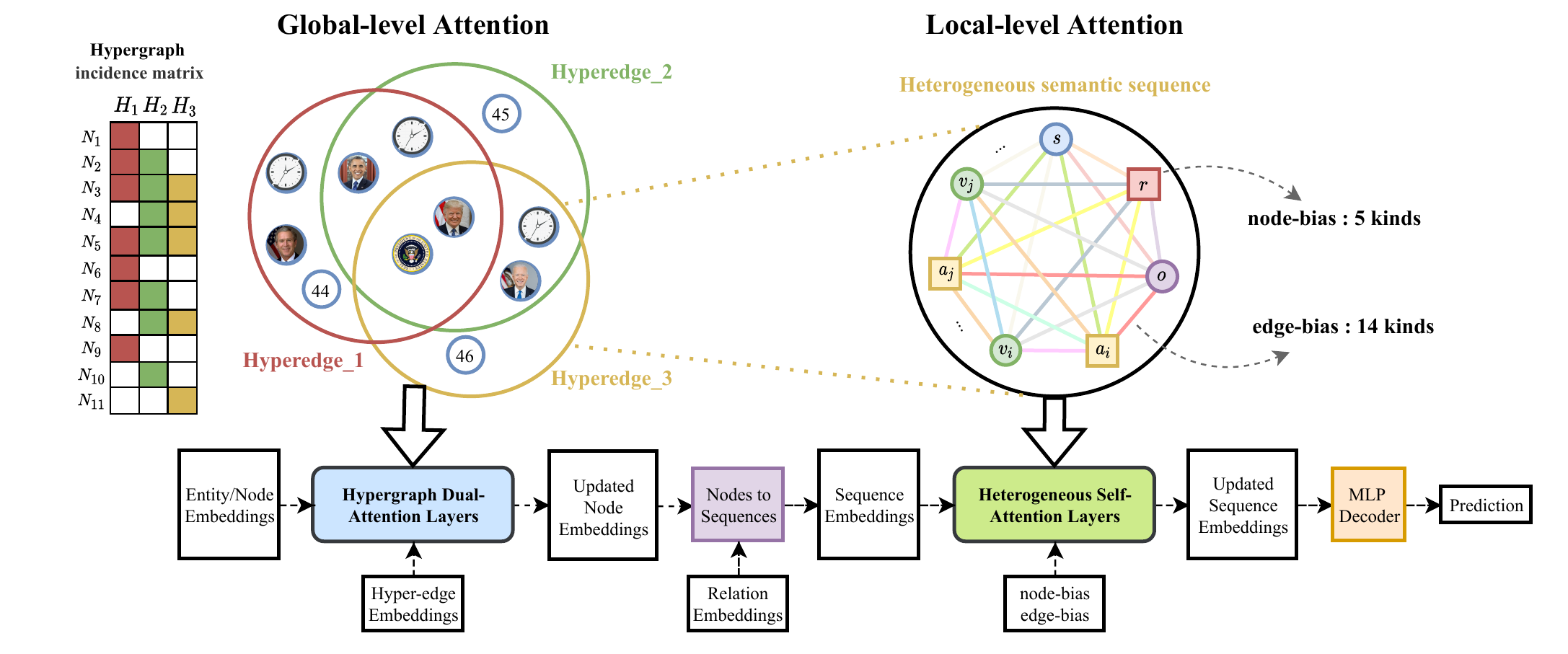}
\caption{The overview of HAHE model for Global-level and Local-level Representation of HKGs.}
\label{HAHE}
\end{figure*}

\section{Preliminaries}

This section presents important concepts and techniques in Hyper-relational Knowledge Graph (HKG), including definitions of HKGs, hypergraph learning, global and local structure of HKGs, and multi-position prediction on HKGs.

\subsection{Hyper-relational Knowledge Graphs} 
HKGs comprise hyper-relational facts (H-Facts). Typically, an H-Fact can be represented as $\mathcal{H}=\{((s,r,o),\{(a_i:v_i)\}^m_{i=1})|s,o,v_1, ...,v_m\in\mathcal{E}, r,a_1,...,a_m\in\mathcal{R}\}$, where $(s,r,o)$ represents the main triple and $\{(a_i:v_i)\}^m_{i=1}$ represents m auxiliary attribute-value qualifiers. 

The link prediction (LP) task on HKGs is to predict missing elements from H-Facts, where missing elements can be entities $\in\{s, o$ $\left., v_{1}, \ldots, v_{m}\right\}$ or relations $\in\left\{r, a_{1}, \ldots, a_{m}\right\}$.

\subsection{Hypergraph Learning on HKGs} 
Since there are more than two entities in an H-Fact, we introduce hypergraph learning~\citep{HGNN}. A hypergraph of HKG, $\mathcal{G}_H=\{\mathcal{E}_H, \mathcal{H_H}, \mathcal{I}_H\}$, contains a node set $\mathcal{E}_H$=$\mathcal{E}$, a hyperedge set $\mathcal{H}_H$=$\mathcal{H}$, and a special incidence matrix $\mathcal{I}_H$ that records the weights of each hyperedge. $\mathcal{I}_H$ is a $|\mathcal{E}_H| \times|\mathcal{H}_H|$ matrix defined as follows:
\begin{equation}
\begin{aligned}
&h(v, e)=1, \text { if } v \in e, \\
&h(v, e)=0, \text { if } v \notin e,
\end{aligned}
\end{equation}
where $v \in \mathcal{E}_H$, $e \in \mathcal{H}_H$. $h$ is a fuction to represent the value in $\mathcal{I}_H$. For a node $v \in \mathcal{E}_H$, its degree is defined as $d(v)=\sum_{e \in \mathcal{H}_H}h(v, e)$, which represents the number of times that a node (entity) appears in different hyperedges (H-Facts) in the whole HKG.

\subsection{Multi-position Prediction on HKGs} 
Multi-position prediction is a new meaningful task with more practicality than the one-position link prediction task on HKGs. For HKG link prediction, in one main triple, we can predict another element for every two of them we know, i.e., we can predict ($s,r,? $), ($s,? ,o$), ($? ,r,o$). In one auxiliary attribute-value qualifier, if we know any of the attributes and values of one of the auxiliary attribute-value qualifiers, we can predict the other, i.e., ($a,? $), ($? ,v$). Thus in practical link prediction, there are some problems have two or more prediction points for example ($s,r,? ,a_1,? ,a_2,v_2$) (prediction position one in the main triple and the other in the first attribute-value qualifier) or ($s,r,o,a_1,? ,a_2,? $) (both predicted positions are in the auxiliary attribute-value qualifiers). We refer to tasks with two or more prediction positions in link prediction tasks as multi-position prediction tasks on HKGs. 

\section{Methodology}
This section introduces our hyper-relational knowledge graph (HKG) embedding model HAHE, including global and local representation, two hierarchical attention layers, and MLP decoder.

\subsection{Global and Local Representation}
HKGs $\mathcal{G}=\{\mathcal{E},\mathcal{R},\mathcal{H}\}$ consist of multiple hyper-relational facts (H-Facts) $\mathcal{H}$ with entities $\mathcal{E}$ and relations $\mathcal{R}$. Since each H-Fact represents an n-ary relation (n>2) and has rich, heterogeneous semantic information, we model HKGs in terms of global-level hypergraph-based representation and local-level sequence-based representation with hypergraph dual-attention layers and heterogeneous self-attention layers respectively. The overview of HAHE is illustrated in Figure~\ref{HAHE}. For global-level representation, we define $\mathcal{G}_H=\{\mathcal{E}_H,\mathcal{H}_H,\mathcal{I}_H\}$ to represent the graph structure of entities. Unlike the regular graph, the hyperedges can connect more than two entity nodes. Moreover, we use incidence matrix $\mathcal{I}_H$ to represent the association information of nodes and hyperedges. For local-level representation, every H-Fact has the structure of one main triple and several auxiliary attribute-value qualifiers $\mathcal{H}=\{((s,r,o),\{(a_i:v_i)\}^m_{i=1})|s,o,v_1, ...,v_m\in\mathcal{E}, r,a_1,...,a_m\in\mathcal{R}\}$, which represents the semantic information of facts. We can fully connect entities and relations in H-Facts and represent them as heterogeneous semantic sequence structure, containing five kinds of nodes $s,r,o,a,v$ and 14 kinds of edges $s-r, s-o, r-o, s-a, s-v, r-a, r-v, o-a, o-v, a_i-a_j, o_i -o_j, a_i-o_i, a_i-o_j$, where $i,j$ are the serial numbers of different qualifiers. 
 
 \begin{figure*}[h!t]
	\centering    %居中
	%\begin{minipage}{0.8 \textwidth}	
	
    \subfigure[Hypergraph Dual-Attention.]{
        \includegraphics[width=0.43\textwidth]{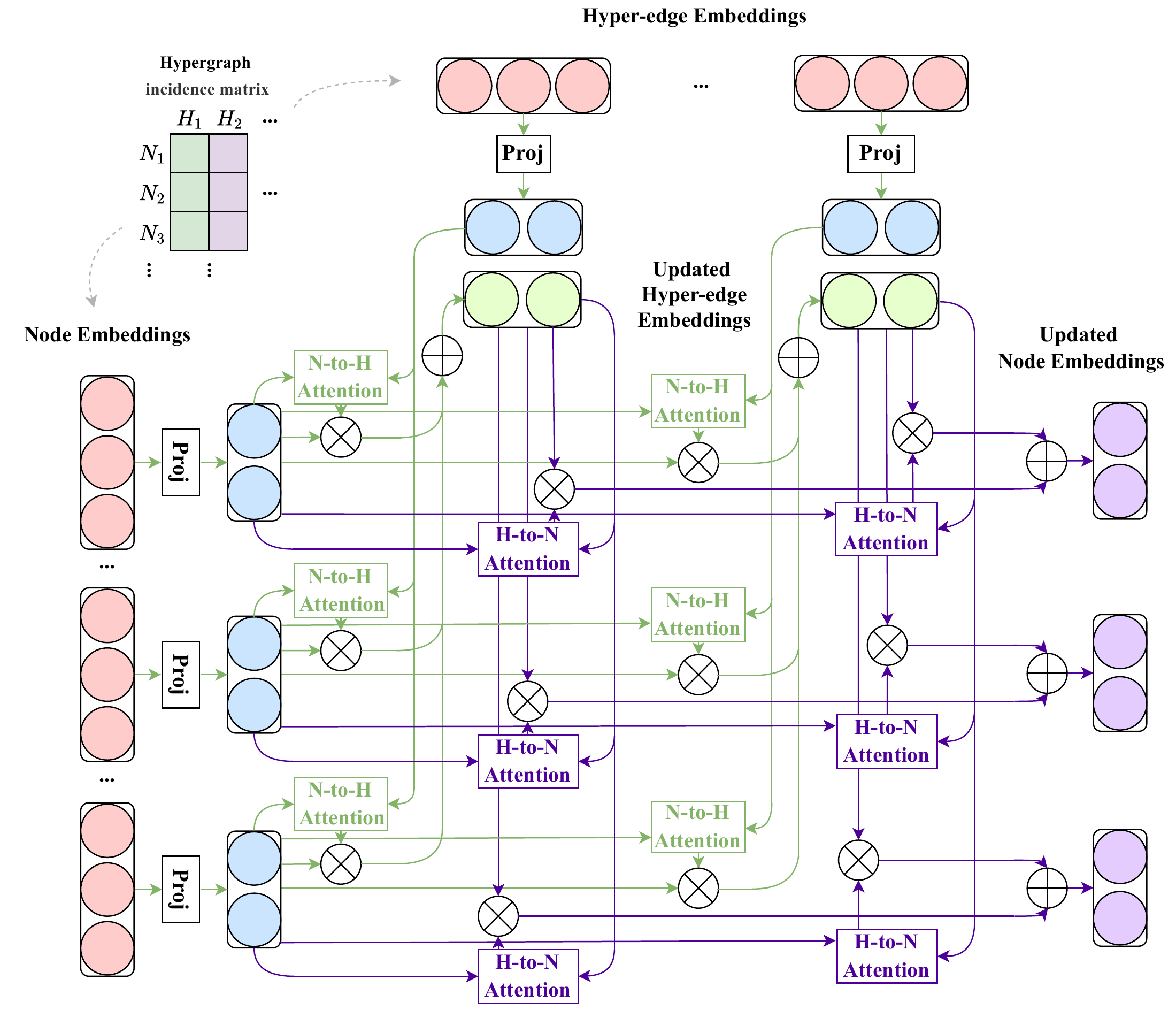}
        \label{41}
    }\hspace{7mm}
    \subfigure[Heterogeneous Self-Attention.]{
        \includegraphics[width=0.43\textwidth]{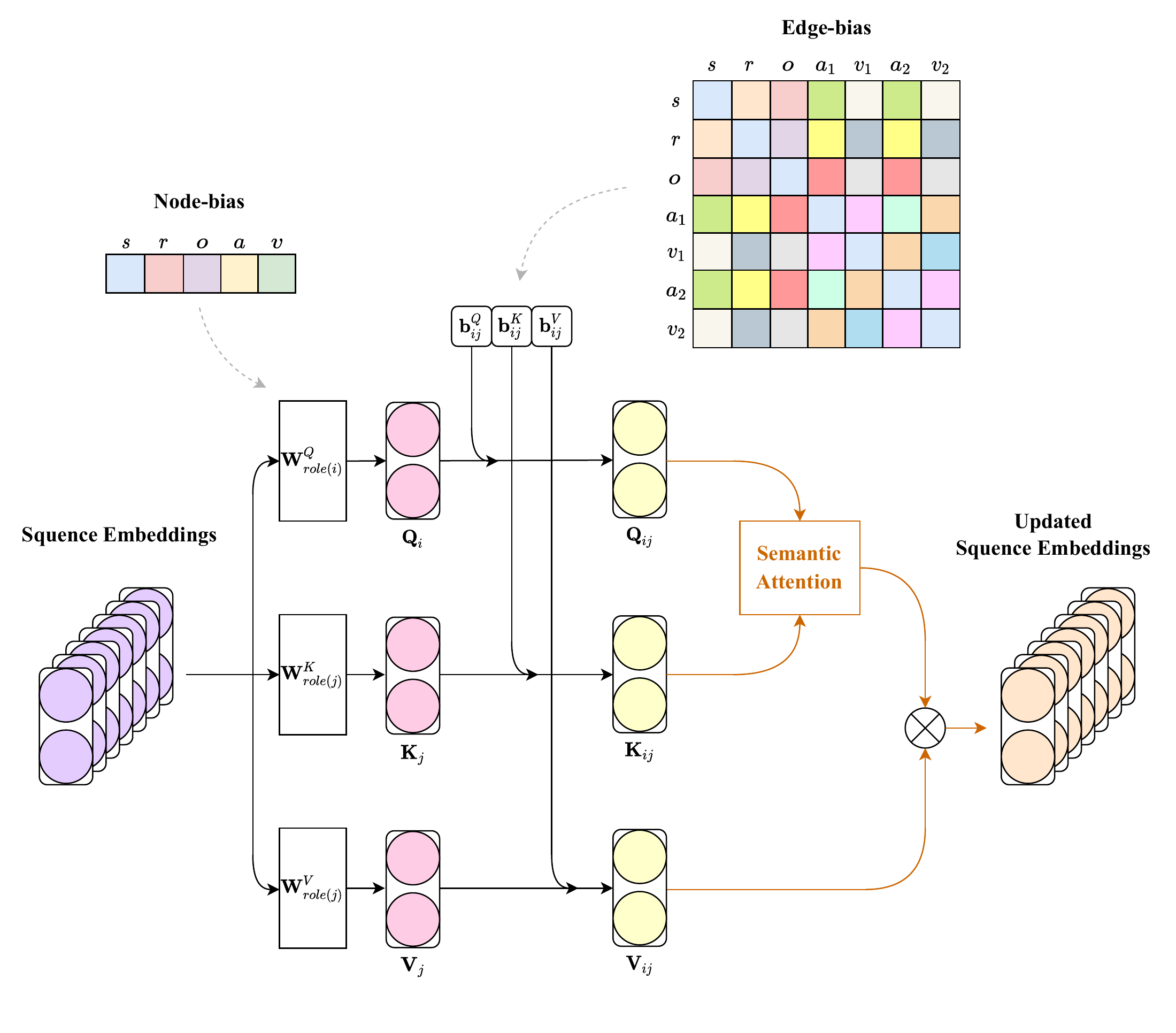}
        \label{42}
    }
	
	\caption{The structure of Hypergraph Dual-Attention Layers and Heterogeneous Self-Attention Layers in HAHE.} %  %大图名称
	\label{fig:1}  %图片引用标记
\end{figure*}

% \subsection{Hierarchical Attention}

% % For the global and local representation of HKG, we propose the hierarchical attention model for HKG embedding, HAHE. It mainly consists of two novel attention layers, Hypergraph Dual-Attention Layers and Heterogeneous Self-Attention Layers. 

\subsection{Hypergraph Dual-Attention Layers}
As shown in Figure~\ref{41}, entity embedding first utilizes Hypergraph Dual-Attention Layers to learn hypergraph structural information in global level. Previous Hypergraph Attention Network methods created a transformed ordinary graph by full joining nodes within the same hyperedge and applying GAT. The hypergraph representation loses because the ordinary graph after this transformation cannot distinguish whether two nodes are within the same hyperedge or different hyperedges. We first initialize the entities as nodes with embedding as $\boldsymbol{h}_{v_i} \in \mathbb{R}^d$, where $v_i \in \mathcal{E}_H$ and $d$ is dimension of embedding, and initialize the H-Facts as hyperedges with embedding as $\boldsymbol{h}_{e_i} \in \mathbb{R}^d$, where $e_i \in \mathcal{H}_H$. Then the node and hyperedge embeddings are projected into the same space to obtain $\mathbf{W} \boldsymbol{h}_{v_i}$ and $\mathbf{W} \boldsymbol{h}_{e_i}$ ,where $\mathbf{W} \in \mathbb{R}^{d \times d}$. The attention from nodes to hyper-edges (N-to-H Attention) is performed as follows:
\begin{equation}
\alpha_{i j}=att\left(\mathbf{W} \boldsymbol{h}_{e_i}, \mathbf{W} \boldsymbol{h}_{v_j}\right) | v_j \in e_i,
\end{equation}
where $\alpha_{i j}$ indicates the importance of node $v_j$’s features to hyperedge $e_i$, $att$ is N-to-H Attention function where we choose a single-layer neural network with concatenation operation. $v_j \in e_i$ means we only calculate the attention where node $v_j$ is in hypergraph $e_i$, which is indexed by hypergraph incidence matrix $\mathcal{I}_H$. Then, the information of nodes is aggregated to hyperedges:
\begin{equation}
\tilde{\boldsymbol{h}}_{e_i}=\sigma\left(\sum_{v_j \in e_i} \frac{\exp\left(\operatorname{LR}\left(\alpha_{i k}\right)\right)}{\sum_{v_k \in e_i} \exp \left(\operatorname{LR}\left(\alpha_{i k}\right)\right)} \mathbf{W} \boldsymbol{h}_{v_j}\right),
\end{equation}
where $\tilde{\boldsymbol{h}}_{e_i} \in \mathbb{R}^d$ is updated hyperedge embeddings and $LR$, $\sigma$ are activation functions. After that, we use a similar way to aggregate the information of updated hyperedges back to nodes with Attention (H-to-N Attention) as follows:
\begin{equation}
\beta_{i j}=att\left(\mathbf{W} \boldsymbol{h}_{v_i}, \mathbf{W} \boldsymbol{h}_{e_j}\right) | e_j \in v_i,
\end{equation}
\begin{equation}
\tilde{\boldsymbol{h}}_{v_i}=\sigma\left(\sum_{e_j \in v_i} \frac{\exp\left(\operatorname{LR}\left(\beta_{i k}\right)\right)}{\sum_{e_k \in {v}_{i}} \exp \left(\operatorname{LR}\left(\beta_{i k}\right)\right)} \tilde{\boldsymbol{h}}_{e_j}\right),
\end{equation}
where $\beta_{i j}$ denotes the importance between updated hyperedge $e_j$ and node $v_i$, and $\tilde{\boldsymbol{h}}_{v_i}$ is the updated node embedding. This way, we update the global hypergraph entity embedding and achieve node-hyperedge-node dual-attention message passing. Though we introduce more hyperedge embeddings than ordinary GNNs, they are only used as an intermediate weight variable for hypergraph attention computation. PyG makes dual-attention easy to implement, making it scalable to large graphs as GNNs.

After hypergraph dual-attention layers, H-Facts distribute updated node embeddings to sequence embeddings with relation embeddings, and fed them into Heterogeneous Self-Attention Layers.

%\begin{equation}
%\tilde{\boldsymbol{h}}_{e_i}=\|_{k=1}^{K}  \sigma\left(\sum_{j \in \aleph_{i}} \alpha_{i j} \mathbf{W} \boldsymbol{h}_{v_j}\right) | v_j \in e_i,
%\end{equation}

\begin{table*}
\small
  \centering
    \begin{tabular}{rrrrrrrrr}
    \toprule
    \multicolumn{1}{c}{Dataset} & \multicolumn{1}{c}{H-Facts} & \multicolumn{1}{c}{H-Facts with Q(\%)} & \multicolumn{1}{c}{Entities} & \multicolumn{1}{c}{Relations} & \multicolumn{1}{c}{Train} & \multicolumn{1}{c}{Valid} & \multicolumn{1}{c}{Test} &
    \multicolumn{1}{c}{Arity} \\
    \midrule
    \multicolumn{1}{c}{JF17K} & \multicolumn{1}{c}{100,947} & \multicolumn{1}{c}{46,320(45.9\%)} & \multicolumn{1}{c}{28,645} & \multicolumn{1}{c}{501} & \multicolumn{1}{c}{76,379} & \multicolumn{1}{c}{-} & \multicolumn{1}{c}{24,568} & 
    \multicolumn{1}{c}{2-6} \\
    \multicolumn{1}{c}{WikiPeople} & \multicolumn{1}{c}{369,866} & \multicolumn{1}{c}{9,482(2.6\%)} & \multicolumn{1}{c}{34,839} & \multicolumn{1}{c}{178} & \multicolumn{1}{c}{294,439} & \multicolumn{1}{c}{37,715} & \multicolumn{1}{c}{37,712} & 
    \multicolumn{1}{c}{2-7} \\
    \multicolumn{1}{c}{WD50K} & \multicolumn{1}{c}{236,507} & \multicolumn{1}{c}{32,167(13.6\%)} & \multicolumn{1}{c}{47,156} & \multicolumn{1}{c}{532} & \multicolumn{1}{c}{166,435} & \multicolumn{1}{c}{23,913} & \multicolumn{1}{c}{46,159} & 
    \multicolumn{1}{c}{2-67}\\
    \bottomrule
    \end{tabular}%
\caption{\label{T}
Dataset statistics, where the columns respectively indicate the number of all H-Facts, H-facts with qualifiers, entities, relations, H-facts in train/valid/test sets, and the range of arity of H-facts.}
\end{table*}%

\subsection{Heterogeneous Self-Attention Layers}
 Each element in the sequence $\boldsymbol{x}_{i} \in \mathbb{R}^{d}$ has five roles, including $s,r,o,a,v$, and a total of 14 kinds of edge with other elements in the sequence  $\boldsymbol{x}_{j} \in \mathbb{R}^{d}$. So, as shown in Figure~\ref{42}, we design a heterogeneous self-attentive layer with both \textbf{node-bias} and \textbf{edge-bias} to learn the local semantic information of the H-Facts. The elements in the sequence pass through this attention layer and update the embedding as follows:
\begin{equation}
\gamma_{i j}=\frac{\left(\mathbf{W}_{role(i)}^{Q} \boldsymbol{x}_{i}+\mathbf{b}_{i j}^{Q}\right)^{\top}\left(\mathbf{W}_{role(j)}^{K} \boldsymbol{x}_{j}+\mathbf{b}_{i j}^{K}\right)}{\sqrt{d}},
\end{equation}
\begin{equation}
\tilde{\boldsymbol{x}_{i}}=\sum_{j=1}^{n} \frac{\exp \left(\gamma_{ik}\right)}{\sum_{k=1}^{n} \exp \left(\gamma_{ik}\right)}\left(\mathbf{W}_{role(j)}^{V} \boldsymbol{x}_{j}+\mathbf{b}_{ij}^{V}\right),
\end{equation}
where $\gamma_{i j}$ is the importance between one element in sequence $\boldsymbol{x}_i$ and another $\boldsymbol{x}_j$, $\mathbf{W}_{role(i)}^{Q}, \mathbf{W}_{role(i)}^{K},\mathbf{W}_{role(i)}^{V} \in \mathbb{R}^{d \times d}$ are the linear weight metrics of query, key, value, and five different kinds of $\boldsymbol{x}_{i}$ pass through different weight metrics indexed by $role$ function as the node-bias, and $\mathbf{b}_{i j}^{Q},  \mathbf{b}_{i j}^{K}, \mathbf{b}_{i j}^{V} \in \mathbb{R}^{d}$ are designed as the edge-bias. Then the sequence embeddings are updated by learning the semantic information inside the H-Facts as $\tilde{\boldsymbol{x}_{i}} \in \mathbb{R}^{d}$.

\subsection{MLP Decoder}
Finally, the updated sequence embeddings are selected for the embedding at the position to be predicted $\tilde{\boldsymbol{x}}_{p}$ with MLP decoder to get the prediction distribution and obtain the link prediction results.
\begin{equation}
\boldsymbol{P}=softmax(\boldsymbol{MLP}(\tilde{\boldsymbol{x}}_{p})\mathbf{E}^\top),
\end{equation}
where $\boldsymbol{MLP}:\mathbb{R}^{d}\rightarrow\mathbb{R}^{d}$, $\mathbf{E}\in \mathbb{R}^{|\mathcal{E}|\times d}$ shares the initial element embedding matrix, $\boldsymbol{P}\in \mathbb{R}^{|\mathcal{E}|}$ is obtained after softmax operation, which denotes the similarity probability of $\tilde{\boldsymbol{x}}_{p}$ with each element in HKG for obtaining the link prediciton answers.

\subsection{Learning Strategy}
\label{LS}

The model trains through the final loss, which is calculated by the similarity between the target of prediction and all entities:
\begin{equation}
\mathcal{L}=\sum_{t=1}^{|\mathcal{E}|}{\mathbf{y}_t\log \boldsymbol{P}},
\end{equation}
where $\mathbf{y}_t$ is the $t$-th entry of the label $\mathbf{y}$. 

% Since each prediction have more than one answer, we follow~\citep{GRAN} and use label smoothing to define the label $\mathbf{y}$. We set $\mathbf{y}_t = 1- \epsilon$ for the target entity and $\mathbf{y}_t=\frac{\epsilon}{|\mathcal{E}|-1}$ for each of the other entities, with label smoothing rate $\epsilon \in \mathbb{R}^{d}$.
For Multi-position Prediction, due to the fully connected attention mechanism, HAHE can accomplish the multi-position prediction task of HKG by masking two or more entities or relations at two or more positions in the same hyper-relational fact. After passing the MLP encoder, HAHE can get the prediction value $\boldsymbol{P}_i(i\geq 2)$ of the corresponding positions, and find the entity or relaiton with the highest similarity among all HKG elements as the prediction results respectively.

% \subsubsection{Attention Mask Strategy for Linear Time Complexity:}
% Since some HKG datasets such as WD50K~\citep{StarE}, or practical applications, the arity number of entities in H-Facts is relatively large, and the attention of full join has $O(n^2)$ time complexity. Inspired by BigBird~\citep{BigBird} to treat three elements in the main triple as global attention, and keep only the attention of the corresponding attributes and values in auxiliary attribute-value qualifiers, and mask out the other attentions. This reduces the time complexity to near linearity $O(n)$ in the task of link prediction on higher-arity H-Facts with almost no loss of results.

\begin{table*}[h!t]
%\small
%\footnotesize
\scriptsize
\centering
\setlength{\tabcolsep}{0.7mm}{
\begin{tabular}[width=0.85\textwidth]{lrrr|rrr|rrr|rrr|rrr|rrr}
\toprule
\multirow{4.5}{*}{\textbf{Model}} & \multicolumn{6}{c}{\textbf{JF17K}} & \multicolumn{6}{c}{\textbf{Wikipeople}} & \multicolumn{6}{c}{\textbf{WD50K}} \\
 \cmidrule(lr){2-7} \cmidrule(lr){8-13} \cmidrule(lr){14-19}
   & \multicolumn{3}{c}{\textbf{subject / object}} & \multicolumn{3}{c}{\textbf{all entities}} & \multicolumn{3}{c}{\textbf{subject / object}} & \multicolumn{3}{c}{\textbf{all entities}} & \multicolumn{3}{c}{\textbf{subject / object}} & \multicolumn{3}{c}{\textbf{all entities}}\\
  \cmidrule(lr){2-4} \cmidrule(lr){5-7} \cmidrule(lr){8-10} \cmidrule(lr){11-13} \cmidrule(lr){14-16} \cmidrule(lr){17-19}
   & \textbf{MRR} & \textbf{H@1} & \textbf{H@10} & \textbf{MRR} & \textbf{H@1} & \textbf{H@10} & \textbf{MRR} & \textbf{H@1} & \textbf{H@10} & \textbf{MRR} & \textbf{H@1} & \textbf{H@10} & \textbf{MRR} & \textbf{H@1} & \textbf{H@10} & \textbf{MRR} & \textbf{H@1} & \textbf{H@10} \\
 \midrule
m-TransH & 0.206 & 0.206 & 0.462 & 0.102 & 0.069 & 0.168 & 0.063 & 0.063 & 0.300 & - & - & - & - & - & - & - & - & - \\
RAE & 0.215 & 0.215 & 0.466 & 0.310 & 0.219 & 0.504 & 0.058 & 0.058 & 0.306 & 0.172 & 0.102 & 0.320 & - & - & - & - & - & - \\
NaLP & 0.221 & 0.165 & 0.331 & 0.366 & 0.290 & 0.516 & 0.408 & 0.331 & 0.546 & 0.338 & 0.272 & 0.466 & - & - & - & 0.224 & 0.158 & 0.330 \\
NeuInfer & 0.449 & 0.361 & 0.624 & 0.473 & 0.397 & 0.618 & 0.476 & 0.415 & 0.585 & 0.333 & 0.259 & 0.477 & 0.243 & 0.176 & 0.377 & 0.228 & 0.162 & 0.341 \\
HINGE & 0.431 & 0.342 & 0.611 & 0.517 & 0.436 & 0.675 & 0.342 & 0.272 & 0.463 & 0.350 & 0.282 & 0.467 & - & - & - & 0.232 & 0.164 & 0.343 \\
StarE &	0.574 &	0.496 &	0.725 &	0.542 &	0.454 &	0.685 &	0.491 &	0.398 &	0.592 &	0.378 &	0.265 &	0.542 &	0.349 &	0.271 &	0.496 &	- &	- &	- \\
Hyper2 & 0.583 & 0.500	& 0.746 & - & - & - & 0.461	& 0.391	& 0.597	& - & - & -	&	- &	- &	- \\
HyTransformer &	0.582 &	0.501 &	0.742 &	- & - & - & 0.501 &	0.426 &	0.634 & - & - & - & 0.356 & 0.281 &	0.498 & - & - & - \\
GRAN &	0.617 &	0.539 &	0.770 &	0.656 &	0.582 &	0.799 &	0.503 &	0.438 &	0.620 &	0.479 &	0.410 &	0.604 & - & - & - & 0.309 &	0.24 &	0.441\\
MSeaHKG & - & - & - & 0.577 & 0.481	& 0.711	& - & - & - & 0.395	& 0.291	& 0.554	& - & - & - & - & - & - \\

 \midrule \midrule
HAHE &	\textbf{0.623} &	\textbf{0.554} &	\textbf{0.806} &	\textbf{0.668} &	\textbf{0.597} &	\textbf{0.816} &	\textbf{0.509} &	\textbf{0.447} &	\textbf{0.639} &	\textbf{0.495} &	\textbf{0.420} &	\textbf{0.631} &	\textbf{0.368} &	\textbf{0.291} &	\textbf{0.516} &	\textbf{0.402} &	\textbf{0.327} &	\textbf{0.546} \\
HAHE w/o global & 0.621	& 0.548	& 0.787	& 0.659	& 0.588	& 0.797	& 0.501	& 0.434	& 0.629	& 0.483	& 0.407	& 0.612	& 0.356	& 0.280	& 0.501	& 0.390	& 0.315	& 0.531 \\
HAHE w/o node-bias	& 0.620	& 0.546	& 0.787	& 0.659	& 0.587	& 0.797	& 0.474	& 0.429	& 0.622	& 0.487	& 0.405	& 0.611	& 0.354	& 0.283	& 0.506	& 0.396	& 0.313	& 0.529\\
HAHE w/o edge-bias	& 0.620	& 0.545	& 0.786	& 0.657	& 0.586	& 0.796	& 0.503	& 0.435	& 0.629	& 0.483	& 0.407	& 0.612	& 0.357	& 0.281	& 0.513	& 0.391	& 0.316	& 0.533\\

\bottomrule                        
\end{tabular}}
\caption{\label{t4}
Comparison of HAHE with other models, composed of entity prediction accuracy on JF17K, WikiPeople and WD50K. Results of the models are mainly taken from the original paper. Best results in each tasks are in \textbf{bold}. 
}
\end{table*}

\section{Experiments}
This section introduces the experimental settings, results and analysis. We answer the following research questions (RQs).
\textbf{RQ1:} Can HAHE outperform other Hyper-relational Knowledge Graphs (HKG) embedding models on HKG datasets?
\textbf{RQ2:} How does the hierarchical attention mechanism contribute to HAHE?
\textbf{RQ3:} How do hypergraph dual-attention mechanisms contribute to HAHE in global level?
\textbf{RQ4:} How do heterogeneity of nodes and edges in hyper-relation fact contribute to HAHE in local level?
\textbf{RQ5:} How HAHE performs in multi-position prediction tasks on HKGs?

\subsection{Experimental Setup}

\subsubsection{Datasets} We conduct experiments on three hyper-relational datasets \textbf{JF17K}~\citep{m-TransH}, \textbf{WikiPeople}~\citep{NaLP}, and \textbf{WD50K}~\citep{StarE}, respectively, as shown in Table \ref{T}. Among them, JF17K is extracted from Freebase~\citep{Freebase}. WikiPeople is obtained by filtering out the statements containing literals in the original WikiPeople dataset, derived from Wikidata~\citep{Wikidata} concerning entities of type human, and WD50K is a high-quality hyper-relational dataset with richer hyper-relational facts with auxiliary attribute-value qualifiers. 

\subsubsection{Baselines} We compare HAHE against a sizable collection of previous hyper-relational approaches namely: (i) m-TransH~\citep{m-TransH} (ii) RAE~\citep{RAE} (iii) NaLP~\citep{NaLP} (iv) NeuInfer~\citep{NeuInfer} (v) HINGE~\citep{HINGE} (vi) StarE~\citep{StarE} (vii) Hyper2~\citep{Hyper2} (viii) HyTransformer~\citep{HyTransformer} (ix) GRAN~\citep{GRAN} (x) MSeaHKG~\citep{MSeaHKG}.

\subsubsection{Ablations} To evaluate the significance of HAHE's three main modules, hypergraph dual-attention mechanism, node heterogeneity, and edge heterogeneity, we obtain 7 simplified model variants by removing any one or two modules from the full model (\textbf{HAHE-node}, \textbf{HAHE-edge}, \textbf{HAHE-node\&edge}, \textbf{HAHE-global}, \textbf{HAHE-global\&node}, \textbf{HAHE-global\&edge}), and the \textbf{basic} variant by removing all three modules.

\begin{figure*}[h!t]
    \centering
    \subfigure[\label{a}]{
        \includegraphics[width=0.23\textwidth, height=0.2\textwidth]{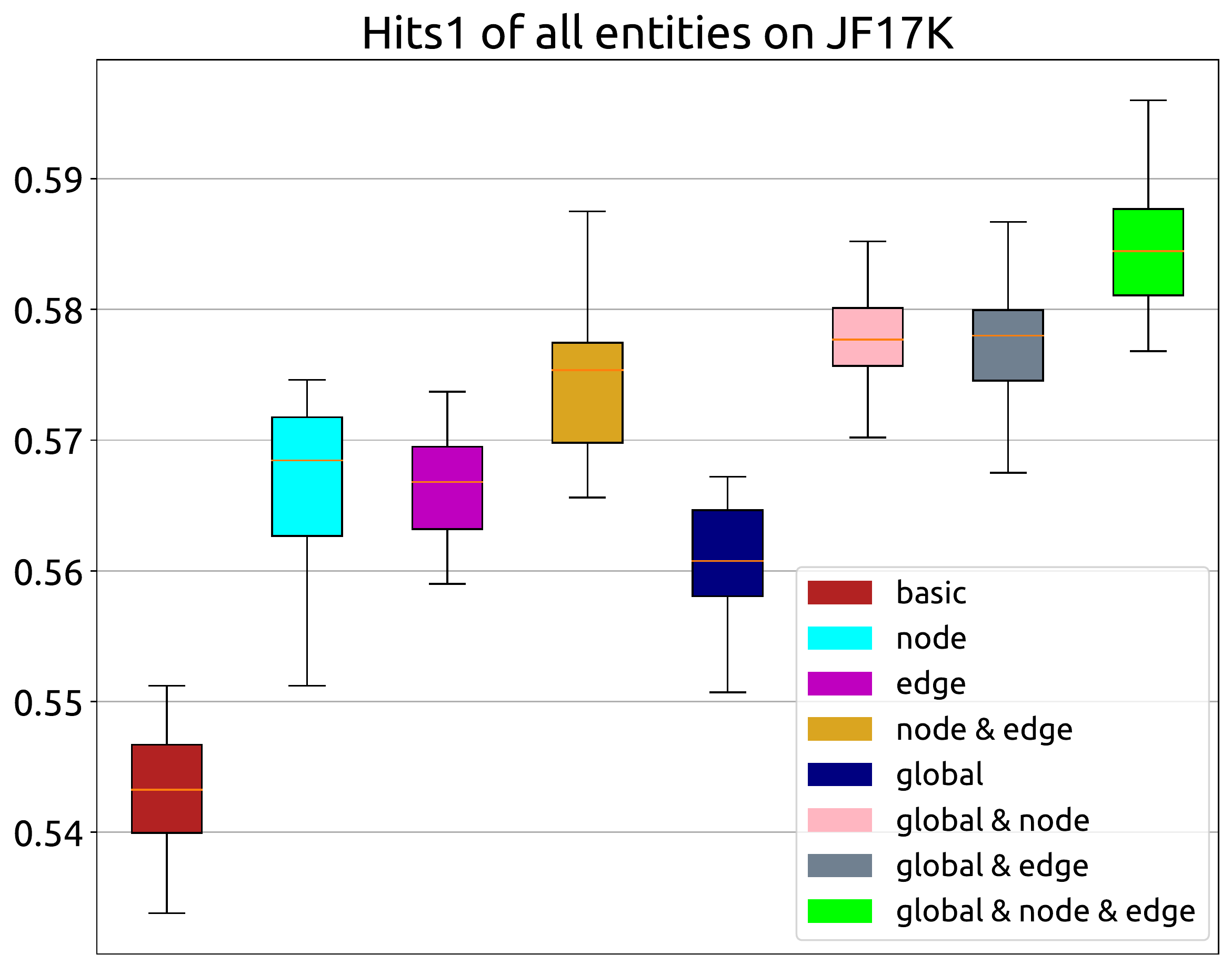}
    }
    \subfigure[\label{b}]{
        \includegraphics[width=0.23\textwidth, height=0.2\textwidth]{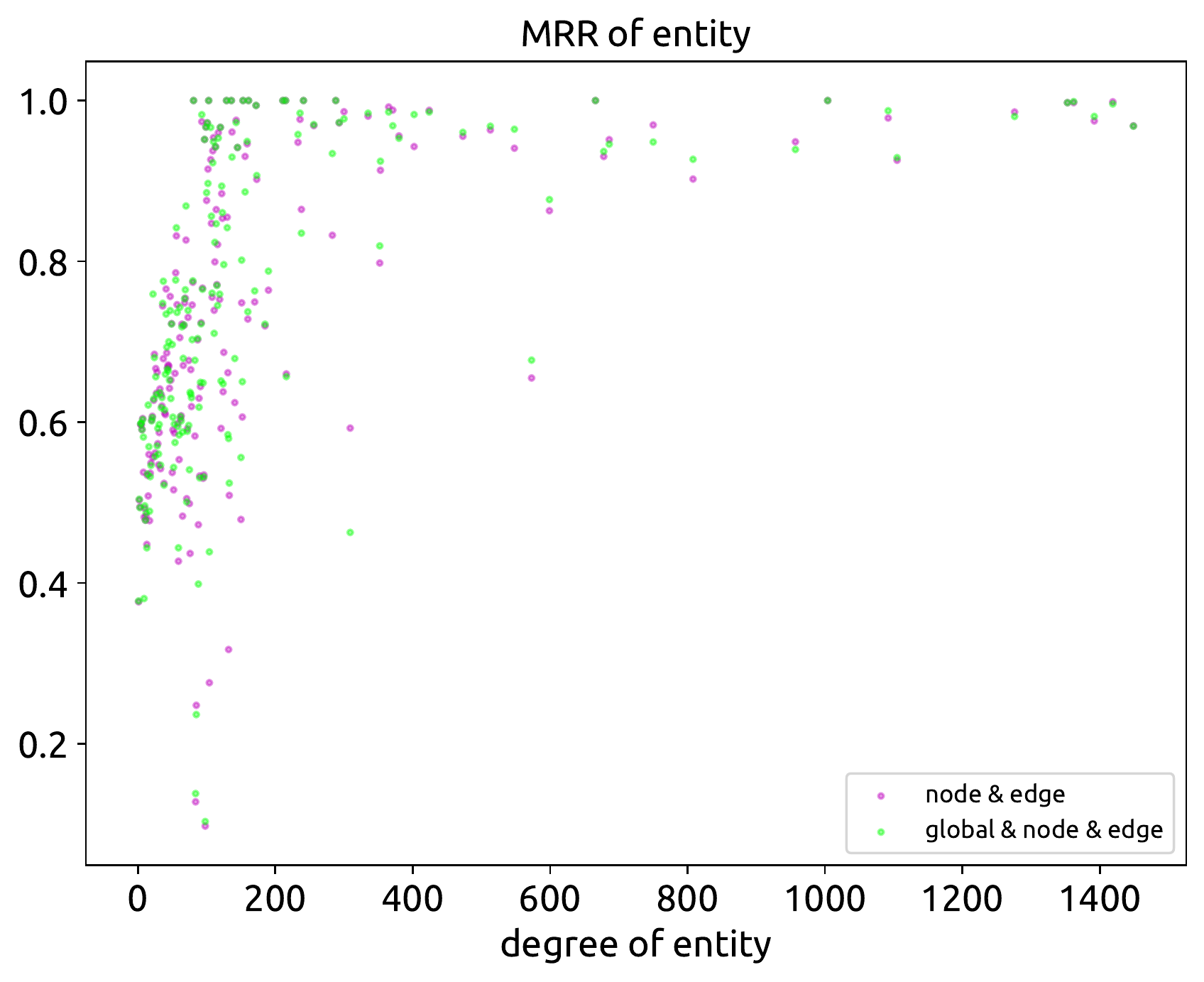}
    }
    %\vfill
    \subfigure[\label{c}]{
        \includegraphics[width=0.23\textwidth, height=0.2\textwidth]{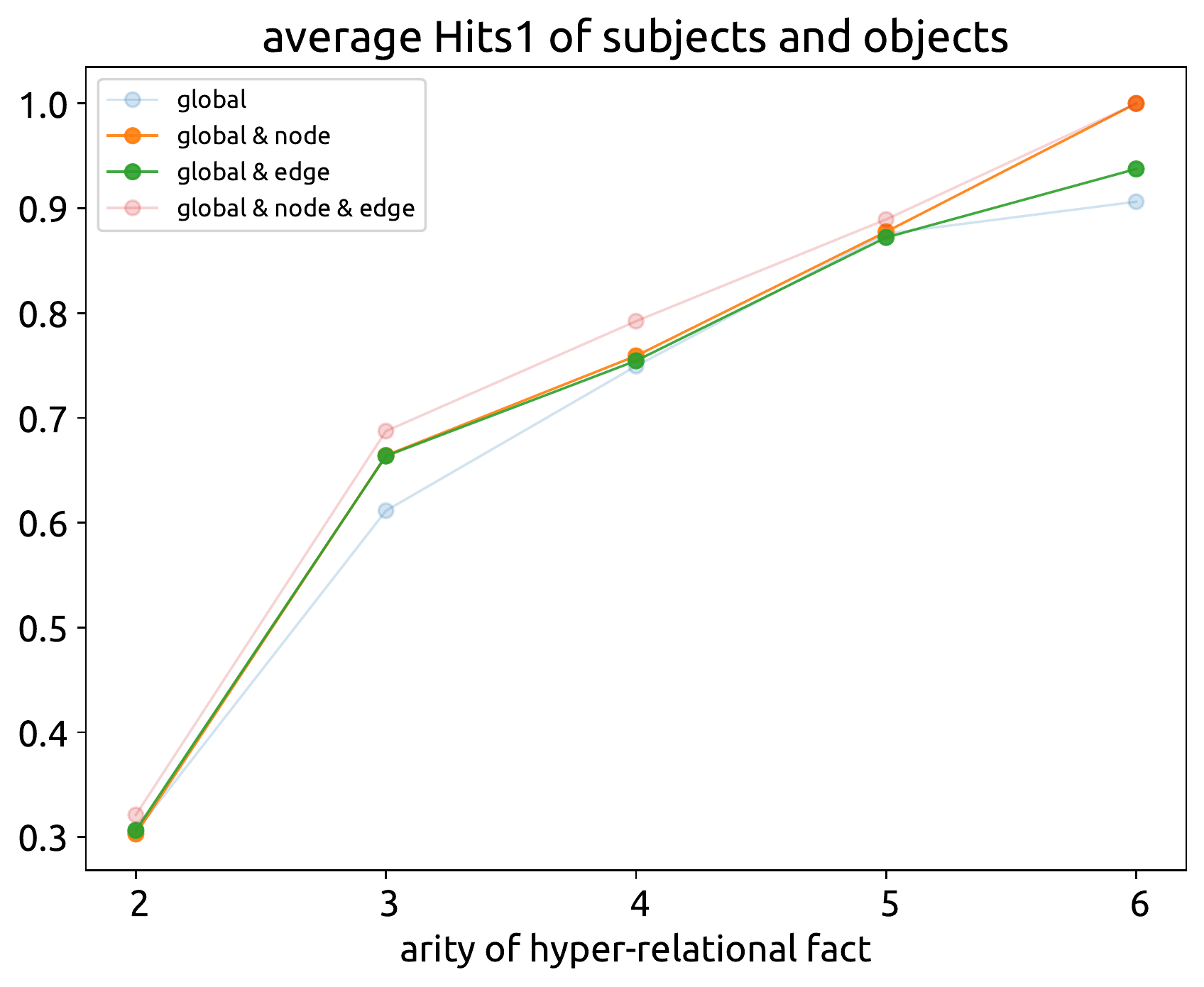}
    }
    \subfigure[\label{d}]{
        \includegraphics[width=0.23\textwidth, height=0.2\textwidth]{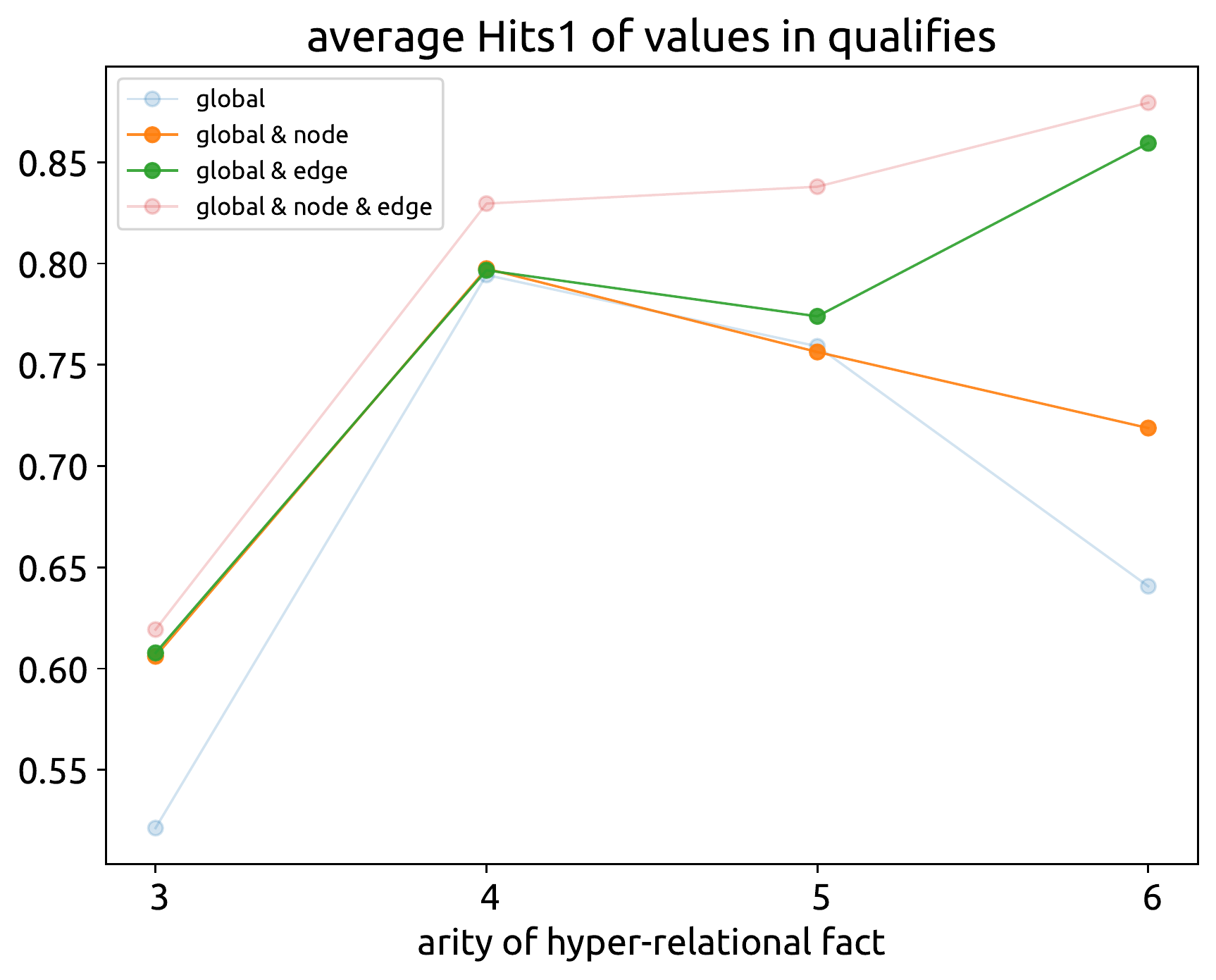}
    }
	\caption{Ablation results. (a) Hits@1 results for ablation study with all HAHE variants. (b) MRR results with different degrees of entity for global-level analysis. (c) Hits@1 results of subject/object prediction with diffent arity for local-level analysis. (d) Hits@1 results of prediction in qualifiers with diffent arity for local-level analysis.}
	\label{f3}
\end{figure*}

\subsubsection{Evaluation Metrics}Each model predicts entities and relations separately. We split each task of predictions into subject/object prediction in main triples and all entities prediction in whole H-facts to test the model's main triple prediction ability. MRR (the average of reciprocal rankings) and Hits@K\ (the proportion of top K\ rankings) for K=1,10 are used to evaluate each link prediction task.

\subsubsection{Hyperparameters and Enviroment}
The model was trained for 300 epochs using the Adam optimizer with a batch size of 1024 examples across 1 GeForce GTX 1080Ti on each dataset. Appendix~\ref{hyper} shows HAHE's optimal hyperparameter settings. Appendix~\ref{train} shows training details.

\subsection{Main Results (RQ1)}

In this experiment, we evaluate our model on the link prediction task. For entity prediction, the results of our model and each variant of our model can be found in Table~\ref{t4}. For relation prediction, the result is shown in Appendix~\ref{RP}. We can observe that the HAHE outperforms the other current methods on all three datasets. On JF17K, for the prediction of subject and object, HAHE reports an improvement of 0.6 (0.9\%) MRR points, 1.5 (2.7\%) H@1, and 3.6 (4.6\%) H@10 compared with the best approach. For the prediction of all entities, HAHE reports a gain of 1.2 (1.8\%) MRR, 1.5 (2.5\%) H@1, and 1.7 (2.1\%) H@10 compared with the next-best approach. For the other two datasets, we also have different degrees of improvement. For WD50K, the latest high-quality HKG dataset, our model has the largest improvement over the existing SOTA model GRAN, with an MRR improvement of about 10 points, which proves that this model is more suitable for hypergraph-structured knowledge graphs with hyper-relational facts beyond binary relation.

\subsection{Ablation Study (RQ2)}

The hypergraph dual-attention mechanism, node heterogeneity, and edge heterogeneity are the three components of HAHE that are required for its operation. We evaluate seven different variants of HAHE, irrespective of whether or not each component is helpful. When evaluating each model variant with a variety of hyperparameters, the results of the optimal prediction were recorded. For different HAHE variants in Figure \ref{a}, it can be observed that hypergraph dual-attention, node heterogeneity, and edge heterogeneity all contribute to the accurate result of our complete model. In addition, we have outlined the specific results of three primary  HAHE variants in Table~\ref{t4}. Each variant lacks a necessary component that is required. Through comparison, our experiment results intuitively demonstrated the effectiveness of HAHE.

Then, we did more refined ablation analysis to explore the significance of the hypergraph dual-attention mechanism in global level and heterogeneity of nodes and edges in local level, respectively.

\subsection{Analysis of Hypergraph Dual-attention Mechanism in Global Level (RQ3)}

We statistically displayed the entity evaluation results of JF17K by the degree of entities in the HKG hypergraph structure to investigate the hypergraph dual attention mechanism. As shown in Figure~\ref{b}, the hypergraph dual-attention mechanism improves the prediction accuracy of entities with different degree. Due to the presence of the attention mechanism, the entity feature information will help in message passing of global information by the hyper-relational facts (hyperedges). For entities with higher degree, the hypergraph dual-attention mechanism can better capture the global features of the hyperrelational facts. Entities with fewer degrees have little impact on capturing global information, but entities with more degrees compensate for this and improve their prediction accuracy.

\subsection{Analysis of Heterogeneity of Nodes and Edges in Local Level (RQ4)}

The experimental results show that both node-bias and edge-bias with heterogeneity can better distinguish different types of entities in an H-fact and different types of relationships between them in local level. Moreover, we find that node-bias plays a more prominent role than edge-bias in the subject/object prediction tasks as shown in Figure \ref{c}, because the entity roles in the main triplet are more diverse than those in the qualifier. And on the element prediction task in qualifiers, edge-bias is better than node-bias as shown in Figure \ref{d}, because edge-bias can distinguish the relationship between corresponding attribute-value pairs $(a_i, v_i)$ and non-corresponding attribute-value pairs $(a_ j, v_i)$ rather than node-bias.

\subsection{Results of Multi-position Prediction Tasks on HKGs (RQ5)}

\begin{figure}[ht]
\centering
\includegraphics[width=7.7cm]{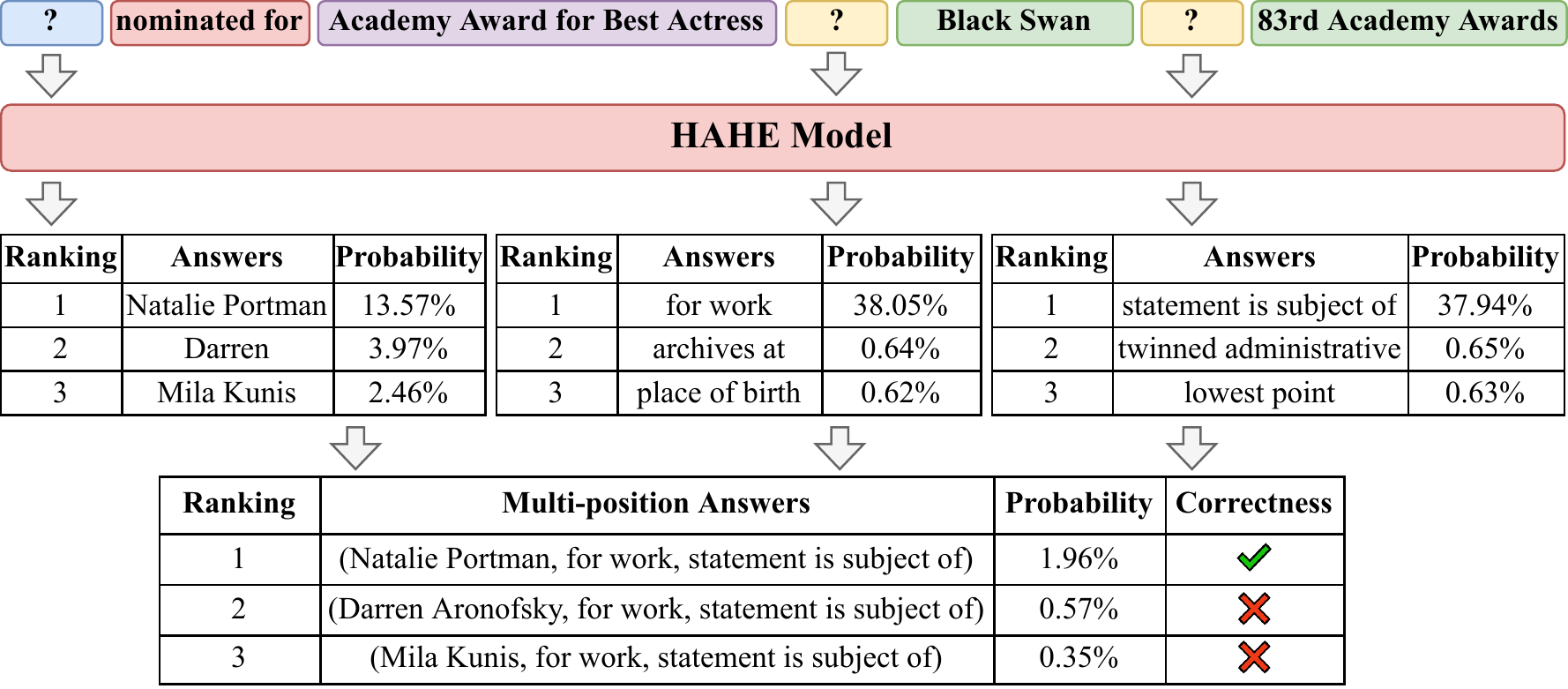}
\caption{Case study of Multi-position Prediction on WD50K.}
\label{case}
\end{figure}

JF17K, WikiPeople, and WD50K were applied to test our multi-position prediction model. As an example, in Figure~\ref{case}, our model outputs the embedding for each position and calculates the joint probability distribution for each candidate answer tuple. Because the increase of predicted positions leads to more answer sets than predicted by the unit placement link, we set the evaluation threshold to keep only the higher scoring answer tuples to obtain the evaluation results. 

\begin{table}[ht]
\scriptsize
%\footnotesize
\centering
\setlength{\tabcolsep}{0.7mm}{
\begin{tabular}{l|ccc|ccc|ccc}
\toprule
\textbf{} & \multicolumn{3}{c|}{\textbf{JF17K}}  &\multicolumn{3}{c|}{\textbf{Wikipeople}}                               & \multicolumn{3}{c}{\textbf{WD50K}}                                                                                           \\ \midrule
\multicolumn{10}{c}{\textbf{2-position prediction}}                                                                                                                                                     \\ \midrule
\textbf{} & \multicolumn{1}{c}{\textbf{av}} & \multicolumn{1}{c}{\textbf{sro/av}} & \multicolumn{1}{c|}{\textbf{all}} & \multicolumn{1}{c}{\textbf{av}} & \multicolumn{1}{c}{\textbf{sro/av}} & \multicolumn{1}{c|}{\textbf{all}} & \multicolumn{1}{c}{\textbf{av}} & \multicolumn{1}{c}{\textbf{sro/av}} & \multicolumn{1}{c}{\textbf{all}} \\ \midrule
\textbf{Ent-Ent}   & 0.039                                   & 0.220                                 & 0.198               & 0.265                                   & 0.129                                 & 0.140                 & 0.326                                   & 0.157                                  & 0.184                                     \\
\textbf{Ent-Rel}   & 0.261                                   & 0.592                                  & 0.540                 & 0.405                                   & 0.409                                  & 0.409                 & 0.527                                   & 0.481                                  & 0.490                                     \\
\textbf{Rel-Rel}   & 0.350                                   & 0.106                                  & 0.159                 & 0.656                                   & 0.544                                  & 0.561                 & 0.940                                   & 0.210                                  & 0.407                                     \\ \midrule 
\multicolumn{10}{c}{\textbf{3-position prediction}}                                                                                                                                                                                                                                                                                                                                                      \\ \midrule 
\textbf{} & \multicolumn{1}{c}{\textbf{av}} & \multicolumn{1}{c}{\textbf{sro/av}} & \multicolumn{1}{c|}{\textbf{all}} & \multicolumn{1}{c}{\textbf{av}} & \multicolumn{1}{c}{\textbf{sro/av}} & \multicolumn{1}{c|}{\textbf{all}} & \multicolumn{1}{c}{\textbf{av}} & \multicolumn{1}{c}{\textbf{sro/av}} & \multicolumn{1}{c}{\textbf{all}} \\ \midrule
\textbf{Ent-Ent}   & 0.011                                   & 0.012                                  & 0.011                 & 0.329                                 & 0.036                                  & 0.046                 & 0.169                                   & 0.350                                  & 0.052                                     \\
\textbf{Ent-Rel}   & 0.022                                   & 0.126                                  & 0.115                 & 0.432                                   & 0.132                                  & 0.144                 & 0.332                                   & 0.166                                  & 0.193                                     \\
\textbf{Rel-Rel}   & 0.029                                   & 0.005                                  & 0.009              & 0.647                                   & 0.162                                  & 0.192                 & 0.884                                   & 0.039                                  & 0.230                                     \\ \bottomrule
\end{tabular}}
	\caption{The MRR results of Multi-position Prediction.}
	\label{tabel_multi}
\end{table}

As shown in Table~\ref{tabel_multi}, we evaluated 2-position prediction and 3-position prediction on three HKG datasets, respectively, and used MRR to rank the answer tuples. The test set has three categories: Ent-Ent, Ent-Rel, and Rel-Rel. Ent-Ent and Rel-Rel indicate that all predicted locations are entities or relations, and Ent-Rel indicates that entities and relations are missing jointly. In addition, whether the main triple is complete divides the sample into two categories. "av" indicates that all predicted positions are in the auxiliary qualifiers, while "sro/av" indicates that the main triple has lost a position" and others in qualifiers. "all" includes both categories. According to the results of multi-position prediction tasks, HAHE performs better on the high-quality WD50K dataset, and more positions predicted or one position in the main triple makes the prediction more difficult. 

\section{Conclusion}
In this paper, we present HAHE, a model with hierarchical attention in global and local level. HAHE outperforms other baselines link prediction tasks on hyper-relational knowledge graphs (HKGs). The experimental results demonstrate that our hypergraph dual-attention layers and heterogeneous self-attention layers are effective in learning the global and local structure of HKGs. We also use HAHE to solve the HKG multi-location prediction task and analyze the results for the first time.

\section*{Acknowledgements}
This work is supported by the National Science Foundation of China (Grant No. 62176026) and Beijing Natural Science Foundation (M22009). This work is also supported by the BUPT Postgraduate Innovation and Entrepreneurship Project led by Haoran Luo.

\section*{Limitations}
For HKG one-position link prediction tasks, HAHE shows the best performance in all three datasets. However, because HAHE is based on hypergraph learning, it improves more on the WD50K high quality hyper-relational knowledge graph link prediction dataset, and less on the Wikipeople dataset where triples are the majority, so HAHE prefers the fact with more arity numbers. In the future, we will consider extending our approach to triples as a unified architecture.

For HKG multi-position link prediction tasks, it can be seen that our model is effective when predicting multiple missing auxiliary information, which is a frequent situation in practical applications. However, the prediction accuracy of our model needs to be further improved in the case of missing primary relations.

\section*{Ethics Statement}
This paper investigates the problem of knowledge graph link prediction, aiming at complementing incomplete hyper-relational knowledge graphs using deep learning methods to better promote knowledge graphs for assisted decision-making and intelligent question-and-answer applications. Therefore, we believe it does not violate any ethics.

% \section*{Acknowledgements}
% This document has been adapted by Jordan Boyd-Graber, Naoaki Okazaki, Anna Rogers from the style files used for earlier ACL, EMNLP and NAACL proceedings, including those for
% EACL 2023 by Isabelle Augenstein and Andreas Vlachos,
% EMNLP 2022 by Yue Zhang, Ryan Cotterell and Lea Frermann,
% ACL 2020 by Steven Bethard, Ryan Cotterell and Rui Yan,
% ACL 2019 by Douwe Kiela and Ivan Vuli\'{c},
% NAACL 2019 by Stephanie Lukin and Alla Roskovskaya, 
% ACL 2018 by Shay Cohen, Kevin Gimpel, and Wei Lu, 
% NAACL 2018 by Margaret Mitchell and Stephanie Lukin,
% Bib\TeX{} suggestions for (NA)ACL 2017/2018 from Jason Eisner,
% ACL 2017 by Dan Gildea and Min-Yen Kan, NAACL 2017 by Margaret Mitchell, 
% ACL 2012 by Maggie Li and Michael White, 
% ACL 2010 by Jing-Shin Chang and Philipp Koehn, 
% ACL 2008 by Johanna D. Moore, Simone Teufel, James Allan, and Sadaoki Furui, 
% ACL 2005 by Hwee Tou Ng and Kemal Oflazer, 
% ACL 2002 by Eugene Charniak and Dekang Lin, 
% and earlier ACL and EACL formats written by several people, including
% John Chen, Henry S. Thompson and Donald Walker.
% Additional elements were taken from the formatting instructions of the \emph{International Joint Conference on Artificial Intelligence} and the \emph{Conference on Computer Vision and Pattern Recognition}.

% Entries for the entire Anthology, followed by custom entries
\bibliography{anthology,custom}
\bibliographystyle{acl_natbib}

\appendix

\newpage
\section*{Appendix} 

\section{Hyperparameter Settings}
\label{hyper}

We use the grid search method to select the optimal hyperparameter settings for the network. The average Hits1 predicted by all entities is chosen as the evaluation metric. The hyperparameters that we can adjust and the possible values of the hyperparameters are first determined according to the structure of our model in Table~\ref{t6}. 

Afterwards, the different hyperparameter choices are combined and the predictive metrics after 50 epochs of training are used to judge the merit of the hyperparameter combinations. The optimal hyperparameter combinations of the model are obtained by circular traversal of all hyperparameter combinations. The optimal hyperparameter combinations are shown in bold. 

% Take hyperparameter choices on JF17K as an example, a 256-dimensional vector is chosen to represent the embedding of entities and relations in the optimal hyperparameter combination. A 2-layer network is chosen for the global part, where attention is calculated using 8 attention heads, and a dropout rate = 0.1 is applied to the embedded representation of the input network. elu is chosen as the activation function between each layer of the global attention.

\begin{table*}[h!t]
%\small
%\footnotesize
\scriptsize
\centering
\begin{tabular}{rrrr}
\toprule
\multicolumn{1}{c}{\textbf{Hyperparameter}} & \multicolumn{1}{c}{\textbf{JF17K}}& \multicolumn{1}{c}{\textbf{Wikipeople}}& \multicolumn{1}{c}{\textbf{WD50K}}\\
\midrule
\multicolumn{1}{c}{Embedding dimension} & \multicolumn{1}{c}{$\left\{128, \textbf{256}, 512, 1024\right\}$}& \multicolumn{1}{c}{$\left\{128, \textbf{256}, 512, 1024\right\}$}& \multicolumn{1}{c}{$\left\{128, \textbf{256}, 512, 1024\right\}$}\\
\multicolumn{1}{c}{Global\underline{\space}layers} & \multicolumn{1}{c}{$\left\{0, 1, \textbf{2}, 3, 4\right\}$}& \multicolumn{1}{c}{$\left\{0, 1, \textbf{2}, 3, 4\right\}$}& \multicolumn{1}{c}{$\left\{0, 1, \textbf{2}, 3, 4\right\}$}\\
\multicolumn{1}{c}{Global dropout} & \multicolumn{1}{c}{$\left\{0.0, 0.1, \textbf{0.2}, 0.3, 0.4, 0.5\right\}$}& \multicolumn{1}{c}{$\left\{0.0, \textbf{0.1}, 0.2, 0.3, 0.4, 0.5\right\}$}& \multicolumn{1}{c}{$\left\{0.0, \textbf{0.1}, 0.2, 0.3, 0.4, 0.5\right\}$}\\
\multicolumn{1}{c}{Global activation} & \multicolumn{1}{c}{$\left\{relu, \textbf{elu}, gelu, tanh\right\}$}& \multicolumn{1}{c}{$\left\{relu, \textbf{elu}, gelu, tanh\right\}$}& \multicolumn{1}{c}{$\left\{relu, \textbf{elu}, gelu, tanh\right\}$}\\
\multicolumn{1}{c}{Global attention heads} & \multicolumn{1}{c}{$\left\{\textbf{4}, 8, 12, 16\right\}$}& \multicolumn{1}{c}{$\left\{\textbf{4}, 8, 12, 16\right\}$}& \multicolumn{1}{c}{$\left\{\textbf{4}, 8, 12, 16\right\}$}\\
\multicolumn{1}{c}{Local\underline{\space}layers} & \multicolumn{1}{c}{$\left\{4, 8, \textbf{12}, 16, 24\right\}$}& \multicolumn{1}{c}{$\left\{4, 8, \textbf{12}, 16, 24\right\}$}& \multicolumn{1}{c}{$\left\{4, 8, \textbf{12}, 16, 24\right\}$}\\
\multicolumn{1}{c}{Local dropout} & \multicolumn{1}{c}{$\left\{0.0, 0.1, \textbf{0.2}, 0.3, 0.4, 0.5\right\}$}& \multicolumn{1}{c}{$\left\{0.0, \textbf{0.1}, 0.2, 0.3, 0.4, 0.5\right\}$}& \multicolumn{1}{c}{$\left\{0.0, \textbf{0.1}, 0.2, 0.3, 0.4, 0.5\right\}$}\\
\multicolumn{1}{c}{Local attention heads} & \multicolumn{1}{c}{$\left\{\textbf{4}, 8, 12, 16\right\}$}& \multicolumn{1}{c}{$\left\{\textbf{4}, 8, 12, 16\right\}$}& \multicolumn{1}{c}{$\left\{\textbf{4}, 8, 12, 16\right\}$}\\
\multicolumn{1}{c}{Decoder activation} & \multicolumn{1}{c}{$\left\{relu, elu, \textbf{gelu}, tanh\right\}$}& \multicolumn{1}{c}{$\left\{relu, elu, \textbf{gelu}, tanh\right\}$}& \multicolumn{1}{c}{$\left\{relu, elu, \textbf{gelu}, tanh\right\}$}\\
\multicolumn{1}{c}{Hidden size} & \multicolumn{1}{c}{$\left\{128, \textbf{256}, 512, 1024, 2048, 4096\right\}$}& \multicolumn{1}{c}{$\left\{128, \textbf{256}, 512, 1024, 2048, 4096\right\}$}& \multicolumn{1}{c}{$\left\{128, \textbf{256}, 512, 1024, 2048, 4096\right\}$}\\
\multicolumn{1}{c}{Batch size} & \multicolumn{1}{c}{$\left\{128, 256, 512, \textbf{1024}, 2046\right\}$}& \multicolumn{1}{c}{$\left\{128, 256, 512, \textbf{1024}, 2046\right\}$}& \multicolumn{1}{c}{$\left\{\textbf{64}, 128, 256, 512, 1024\right\}$}\\
\multicolumn{1}{c}{Learning rate} & \multicolumn{1}{c}{$\left\{0.0001, \textbf{0.0005}, 0.001\right\}$}& \multicolumn{1}{c}{$\left\{0.0001, \textbf{0.0005}, 0.001\right\}$}& \multicolumn{1}{c}{$\left\{0.0001, \textbf{0.0005}, 0.001\right\}$}\\
\multicolumn{1}{c}{Weight decay} & \multicolumn{1}{c}{$\left\{\textbf{0.01}, 0.02\right\}$}& \multicolumn{1}{c}{$\left\{\textbf{0.01}, 0.02\right\}$}& \multicolumn{1}{c}{$\left\{\textbf{0.01}, 0.02\right\}$}\\
\multicolumn{1}{c}{Soft label for entity} & \multicolumn{1}{c}{$\left\{0.5, 0.6, 0.7, 0.8, \textbf{0.9}\right\}$}& \multicolumn{1}{c}{$\left\{0.0, 0.1, \textbf{0.2}, 0.3, 0.4\right\}$}& \multicolumn{1}{c}{$\left\{0.0, 0.1, \textbf{0.2}, 0.3, 0.4\right\}$}\\
\multicolumn{1}{c}{Soft label for relation} & \multicolumn{1}{c}{$\left\{\textbf{0.0}, 0.3, 0.6, 0.9\right\}$}& \multicolumn{1}{c}{$\left\{0.0, \textbf{0.1}, 0.2, 0.3\right\}$}& \multicolumn{1}{c}{$\left\{0.0, \textbf{0.1}, 0.2, 0.3\right\}$}\\
\bottomrule   
\end{tabular}

\caption{\label{t6}
Hyperparameter Search.
}

\end{table*}

\begin{table*}[h!t]
%\small
%\footnotesize
\scriptsize
\centering
\setlength{\tabcolsep}{0.7mm}{
\begin{tabular}[width=0.85\textwidth]{lrrr|rrr|rrr|rrr|rrr|rrr}
\toprule
\multirow{4.5}{*}{\textbf{Model}} & \multicolumn{6}{c}{\textbf{JF17K}} & \multicolumn{6}{c}{\textbf{Wikipeople}} & \multicolumn{6}{c}{\textbf{WD50K}} \\
 \cmidrule(lr){2-7} \cmidrule(lr){8-13} \cmidrule(lr){14-19}
   & \multicolumn{3}{c}{\textbf{main relation}} & \multicolumn{3}{c}{\textbf{all relations}} & \multicolumn{3}{c}{\textbf{main relation}} & \multicolumn{3}{c}{\textbf{all relations}} & \multicolumn{3}{c}{\textbf{main relation}} & \multicolumn{3}{c}{\textbf{all relations}}\\
  \cmidrule(lr){2-4} \cmidrule(lr){5-7} \cmidrule(lr){8-10} \cmidrule(lr){11-13} \cmidrule(lr){14-16} \cmidrule(lr){17-19}
   & \textbf{MRR} & \textbf{H@1} & \textbf{H@10} & \textbf{MRR} & \textbf{H@1} & \textbf{H@10} & \textbf{MRR} & \textbf{H@1} & \textbf{H@10} & \textbf{MRR} & \textbf{H@1} & \textbf{H@10} & \textbf{MRR} & \textbf{H@1} & \textbf{H@10} & \textbf{MRR} & \textbf{H@1} & \textbf{H@10} \\
 \midrule
m-TransH & - & - & - & - & - & - & - & - & -  & - & - & - & - & - & - & - & - & - \\
RAE & - & - & - & - & - & - & - & - & - & - & - & -  & - & - & - & - & - & - \\
NaLp & 0.639 & 0.547 & 0.822 & 0.825 & 0.762 & 0.927 & 0.482 & 0.320 & 0.482 & 0.735 & 0.595 & 0.938 & - & - & - & - & - & -  \\
NeuInfer & 0.936 & 0.901 & 0.989 & - & - & - & 0.950 & 0.915 & \textbf{0.997} & - & - & - & - & - & - & - & - & - \\
HINGE & - & - & - & 0.861 & 0.832 & 0.910 & - & - & - & 0.765 & 0.686 & 0.900 & - & - & - & - & - & - \\
StarE & - & - & - & 0.901 & 0.884 & 0.963 & - & - & - &	0.378 &	0.265 &	0.542 & - & - & - &	- &	- &	- \\
Hyper2 & 0.950 & 0.933	& 0.976 & - & - & - & 0.947	& 0.914	& 0.987	& - & - & -	&	- &	- &	- \\
HyTransformer & - & - & - & - & - & - & - & - & - & - & - & -  & - & - & - & - & - & -\\
GRAN &	0.992 &	0.988 &	\textbf{0.988} &	\textbf{0.996} &	0.993 &	\textbf{0.999} &	\textbf{0.957} &	\textbf{0.942} &	0.976 &	\textbf{0.960} &	\textbf{0.946} &	0.977 & - & - & - & - & - & -\\
MSeaHKG & - & - & - & 0.933 & 0.894	& 0.972	& - & - & - & 0.831	& 0.787	& 0.972	& - & - & - & - & - & - \\

 \midrule \midrule
HAHE &\textbf{ 0.993} & \textbf{0.989} &	\textbf{0.998} &	\textbf{0.996} &	\textbf{0.994} & \textbf{0.999} &	\textbf{0.957} &	0.941 &	0.978 &	0.958 &	0.942 &	\textbf{0.978}  & \textbf{0.916} & \textbf{0.885} & \textbf{0.964} & \textbf{0.927} & \textbf{0.900} & \textbf{0.969} \\
HAHE w/o global	& 0.992 & 0.988 & 0.997	& \textbf{0.996}	& \textbf{0.994}	& 0.998	& 0.946	& 0.930	& 0.967	& 0.947	& 0.931	& 0.968	& 0.915	& 0.884	& 0.961	& \textbf{0.927}	& \textbf{0.900}	& 0.966 \\
HAHE w/o node-bias & 0.989	& 0.984	& 0.995	& 0.994	& 0.991	& 0.997	& 0.926	& 0.906	& 0.952	& 0.926	& 0.906	& 0.953	&	0.915 &	0.883 &	0.963	& 0.922	& 0.899	& 0.969 \\
HAHE w/o edge-bias	& 0.992	& 0.987	& \textbf{0.998}	& 0.995	& 0.993	& \textbf{0.999}	& 0.948	& 0.932	& 0.970	& 0.949	& 0.933	& 0.970	& 0.915	& 0.882	& 0.963	& 0.926	& 0.898	& 0.968 \\

\bottomrule                        
\end{tabular}}
\caption{\label{t5}
Comparison of HAHE with other models, composed of relation prediction accuracy on JF17K, WikiPeople and WD50K. Results of the models are mainly taken from the original paper. Best results in each tasks are in \textbf{bold}. 
}
\end{table*}
% The local attention is a 12-layer network. The data is fed into the local encoder with a dropout rate of 0.2, and the local encoder uses 4 attention heads to compute the attention. After the attention calculation, the vector is projected into a 512-dimensional space, which is transformed by the fully connected layer and finally restored to 256 dimensions. This part uses gelu as the activation function between each fully connected layer. Cross entropy is used as the basis of the loss function, and the soft label values in the label smoothing are adjusted for different data sets to obtain the corrected loss function as the final loss function. The optimal model use Adam with learning rate = 0.0005 as our optimizer.

\section{Model Training Details}
\label{train}

We train 300 epochs in each dataset with the optimal combination of hyperparameters. The prediction results of the network are evaluated using the test sets with the best model. HAHE and all its variants have been trained on a single 11G 1080Ti GPU. Using our optimal hyperparameter settings, the time required to complete the training on the three datasets JF17K, Wikipeople, and WD50K is 5h, 12h and 8h, respectively.

\section{Results of Relation Prediction}
\label{RP}
As shown in Table~\ref{t5}, the result of relation prediction of previous models for HKG embedding has achieved very good results in relation prediction, some indicators have even reached 99\%.

\end{document}